# How to Discover a Semantic Web Service by Knowing Its Functionality Parameters

Golsa Heidary[1], Kamran Zamanifar[2]

[1]Young researchers club,
[2]Dept. of Computer Science,
[1,2]Islamic Azad University,
[1,2]Najafabad Branch, Isfahan, Iran.
[1] golsa_heidary@sco.iaun.ac.ir,
[2] zamanifar@eng.ui.ac.ir

Naser Nematbakhsh[3], Farhad Mardukhi[4]

[3,4]Dept. of Computer Science,
[3,4]University of Isfahan,
[3,4]Isfahan, Iran.
[3] nemat@eng.ui.ac.ir3,
[4] mardukhi@eng.ui.ac.ir4

*Abstract*-In this work, we show how to discover a semantic web service among a repository of web services. A new approach for web service discovering based on calculating the functions similarity. We define the Web service functions with Ontology Web Language (OWL). We wrote some rules for comparing two web services` parameters. Our algorithm compares the parameters of two web services` inputs/outputs by making a bipartite graph. We compute the similarity rate by using the Ford–Fulkerson algorithm. Higher the similarity, the less are the differences between their functions. At last our algorithm chooses the service which has the highest similarity. As a consequence, our method is useful when we need to find a web service suitable to replace an existing one that has failed. Especially in autonomic systems, this situation is very common and important since we need to ensure availability of the application which is based on the failed web service. We use Universal Description, Discovery and Integration (UDDI) compliant web service registry.

## I. INTRODUCTION

Semantic web service discovery is the process of finding an existing Web services based on their functional and nonfunctional semantic description. Discovery scenarios typically occur when we try to reuse an existing Web service in building new or enhanced business processes.

Web Service Description Language (WSDL) gives us a model and an XML format to describe Web services. It shows just the description of the abstract functionality, of a service, but not details of a service description such as "how" and "where". WSDL describes only the syntactic interface of web services and doesn`t show the semantics of a service. Hence, the WSDL cannot be used for selecting semantic web matching.

These standards require a lot of effort for creating and managing end-to-end service interactions. In our dynamic world, where services are created, updated, deleted on the fly, these industry standards are not sufficient. The academic solution to give meaning to web services, so that besides humans, software agents can also have benefit, by means of semantic web issues, that is annotating web services with OWL which is an Ontology Web Language(OWL). This method presents a way for giving an abstract representation of web services which can be characterized with functional properties like input/output (IO). Our method is mainly concerned with the functional properties, i.e. IO of services. It focuses on the capability matching: the matching between IOs of advertised services and IOs of requested services.

This paper is extension of our last work, "A new approach for semantic web matching", and proposes a web service selection framework which uses bipartite graph for computing the similarity Rate of two web services. OWL is used for semantically describing web services. The Ford-Fulkerson algorithm can compute the matching rate between two web services. We show that our proposed framework is capable of selecting most similar web service according to the ontological grading of inputs/outputs, among web services in the repository.

In the remainder of this paper, first in Section II, we talk about related works, in brief. These works are about matching, discovering and composing web services that have used different methods. Then in section III we define the main idea of discovering semantic web services this section has three parts. In the first part we describe the OWL language and the usage of that in our work. Second part is about the computing the similarity rate between two semantic web services and third part is about discovering the most similar web service. Section IV compares different algorithms with each other and shows that our chosen algorithm is the best one and has the least running time among all. Finally in section V the conclusion of our work and future works, are given.

## II. RELATED WORK

Different calculating methods for computing the similarity between Web services are studied and applied in many aspects.

Some of works give the pre-condition and post-condition of services, so that they established a state-chart to carry out the interaction between services [1].

A technique for similarity searching for Web services was proposed by [3]. In this work, they use association-rule-mining approach and a hierarchical clustering algorithm for parameter names of Web services. In [5] by mining the historical invocations of component services, they construct a semantic model to describe their behavior rules based on the qualitative probabilistic network. Works which have groundings probability will not have exact answers in real world. In [4] they gave an approach by modeling the behaviors of a service and it was effective for service composition. In [6] WSDL is the language that they have chosen for describing semantics of web services. As we know, WSDL doesn`t have enough information about semantics. In [8] they introduce a new ontology by the help of genetic algorithm. In [9] they use the ontology based modeling to make semantic models as conceptual frameworks for the semantic description of Web services, in which the ontology is regarded as the semantic annotations. In [10] semantic Web Service Description Ontology which is called SWSD and QOS(Quality Of Service), are used to do matching, which are not as good as functionality matching. In this work also, comparing service functionalities are considered in brief. The similarity measured in [13], is based on statistics, but our work is precise, not on probability. In [11] they use WS-policy for matching web services. We know that comparing functionalities will give a better result for semantic web matching. In this work, by the help of automaton based modeling, they design, specify and verify global behavior of services. Reference [12] matches two services in brief. The similarity rate just considers whether two services are similar, but does not explore how much similar they are and it is more about service composition. And finally, our last work on semantic web matching [2] used bipartite graph for comparing two web services functionalities to find the most similar one. But matching rate was computing with fuzzy methods and was not accurate. Now in this paper, our computations are very precise, because we give a numeric value for matching rate of two semantic web services.

## III. DISCOVERING A WEB SERVICE

We have a repository of semantic web services by the name of UDDI, where all services are described by the OWL language. When we want to select a web service, we should know what it does. It means that by comparing two web services` functionalities, we can compute the similarity rate of two web services. We do this work, for input and output parameters. At last we select the most similar web service among them. So the steps of our work are:

- Making the UDDI repository by Describing services in OWL language
- Computing the similarity rate between each advertised web service and the requested one.
- Selecting the most similar web service

Now we describe all of these steps.

### A. Describing a Service With OWL Language

Our method focuses on semantic services that are described in OWL. In the following, we briefly introduce the essentials of OWL.

#### 1) Overlook

OWL is an upper ontology used to describe the semantics of services based on the W3C standards and is grounded in WSDL. The OWL ontology has three main components:

- The service profile advertises and discovers services;
- The process model gives a detailed description of a service's operation; and the
- Grounding which provides details on how to interoperate with a service, via messages.

In particular, the semantic service profile in OWL specifies the semantics of the service signature that is the inputs required by the service and the outputs generated [5]. Furthermore, because a service may need external conditions to be satisfied, the profile also describes the preconditions to be satisfied before, and after execution of the service, some expected effects may happen. Most of existing OWL service matchmakers focuses on semantic service profiles.

#### 2) OWL Service Profile

The OWL profile ontology is used for the purpose of service discovery and it describes what the service does. An OWL service profile encompasses its functional parameters, i.e. has Output, precondition and effect, and also it encompasses non-functional parameters such as service Name, service Category, quality Rating, text Description, and meta-data about the service provider such as name and location.

#### 3) OWL Service Process Model

An OWL process model describes about composition of one or more subservices of a service, and making a new one. In other words, it is about choreography and orchestration.

Originally, it was not intended for service discovery by the so-called OWL coalition that is the group of researchers who developed OWL.

#### 4) OWL Service Grounding

The grounding of an OWL service description, for facilitating service execution, provides a binding between the logic-based semantic service profile, the process model, and the XML-based Web service interface. Such a grounding of OWL services can be, in principle, arbitrary but has been exemplified for grounding in WSDL (Web Service Description Language) to concretely connect OWL to an existing Web service standard. In particular, the logic-based description of the service signature is uniquely associated with that of the Web service, and an atomic semantic process model is mapped to a WSDL operation.

### B. Computing The Similarity Rate of Each Advertised Web Service with The Requested One

This step has three phases; at first we should apply the rules and then we should make a bipartite graph for both input and output parameters, separately. At last compute the similarity rate by the help of Ford-Fulkerson algorithm.

#### 1) Similarity Function (SIM)

The aim of the SIM function is to calculate the similarity rate between parameters with respect to the closeness of such parameters in a given ontology. In particular, with assumption that the registry includes Web services related to a given application domain, for computing the similarity rate between two parameters, we can rely on both domain-specific ontology and general purpose ontology. The domain-specific ontology includes terms related to a given application domain. The general-purpose ontology includes all the possible terms. We decided to rely on domain-specific ontology since it offers more accuracy in the relationships of the terms. We assume that in a domain-specific ontology each word has a unique sense with respect to the domain itself. We have summarized the rule of comparing two parameters of two semantic web services in table 1. So, according to our rules we can draw the edges of our bipartite graph and give weight to them. These weights are useful for computing the similarity rate. As you see, the weight is between 0 and 10. 0 means that discovery is failed and 10 means that two services match completely.

TABLE I. RULES OF COMPARING TWO PARAMETERS OF TWO WEB SERVICES

| | | Q Parameter | | | | |
|---|---|---|---|---|---|---|
| | Data Type | Integer | Real | String | Date | Boolean |
| R Parameter | Integer | 10 | 5 | 3 | 1 | 1 |
| | Real | 10 | 10 | 1 | 0 | 1 |
| | String | 7 | 7 | 10 | 8 | 3 |
| | Date | 1 | 0 | 1 | 10 | 0 |
| | Boolean | 1 | 0 | 1 | 0 | 10 |

### 2) Making Bipartite Graph

A Bipartite Graph is a graph G = (V, E) in which the vertices set can be partitioned into two disjoint sets, $V = V_0 + V_1$, such that every edge e in E has one vertex in $V_0$ and another in $V_1$.

The matching is complete if and only if, all vertices in $V_0$ are matched. It means that all vertices in $V_0$, as well as $V_1$, should have an edge.

If R is the requested service and Q is an advertised service in UDDI, let $R_{out}$ and $Q_{out}$ be the set of output concepts in R and Q respectively. These constitute the two vertex sets of our bipartite graph. Construct graph G = ($V_0$ + $V_1$, E), where, $V_0$ = $R_{out}$ and $V_1$ = $Q_{out}$ Consider two concepts a in $V_0$ and b in $V_1$. It means that a is one of the output parameters of R and b is one of the output parameters of Q.

### 3) Computing The Similarity Rate

In this part, at first we describe the maximum flow problem and name some algorithms for solving it. Then we show how this problem will help to our selecting method [14].

#### a) Maximum Flow

We should add the source and sink vertices to our bipartite matching graph. In the maximum-flow problem, we will to compute the greatest rate at which material ships from the source to the sink without violating any capacity constraints. This problem can be solved by efficient algorithms. There are two general methods for solving the maximum-flow problem which are Ford-Fulkerson and Edmonds-Karp, and some others which are Push-Relabel, Relabel-to-Front bipartite matching algorithm. For finding the matching rate of two semantic web services, we use the first method. So we describe it in the following.

#### b) Flow networks and flows

A flow network G = (V, E) is a directed graph in which each edge (u, v) ∈ E has a nonnegative capacity $c (u, v)$

≥ 0, Otherwise c (u, v) = 0. We have two vertices in a flow network: a source s and a sink t. For convenience, we assume that every vertex lies on some path from the source to the sink. That is, for every vertex v ∈ V, there is a path s; v; t. So the graph is connected and |E| ≥ |V| − 1.

Each flow network has three properties:

- Capacity constraint:
  $\forall u, v \in V: f(u,v) \leq c(u,v)$
- Skew symmetry:
  $\forall u, v \in V: f(u,v) = -f(v,u)$
- Flow conservation:
  $\forall u \in V - \{s,t\}: \sum_{v \in V} f(u,v) = 0$

#### c) Residual networks

Given a flow network and a flow, the residual network consists of edges that can admit more flow.

Suppose that we have a flow network G = (V, E) with source s and sink t. Let f be a flow in G, and consider a pair of vertices u, v ∈ V. The amount of additional flow we can push from u to v before exceeding the capacity c (u, v) is the residual capacity of (u, v), given by

$$c_f(u, v) = c(u, v) - f(u,v) \qquad (1)$$
$$E_f = \{(u, v) \in V \times V : c_f(u, v) > 0\} \qquad (2)$$

#### d) Augmenting Paths

In a flow network G = (V, E) and a flow f, an augmenting path $p$ is a simple path from s to t in the residual network $G_f$. By the definition of the residual network, each edge (u, v) on an augmenting path admits some additional positive flow from u to v without violating the capacity constraint on the edge [14].

We call the maximum amount by which we can increase the flow on each edge in an augmenting path p, the residual capacity of p, given by cf(p) :

$$c_f(p) = min \{c_f(u, v) : (u, v) \text{ is on } p\} \qquad (3)$$

#### e) The Ford-Fulkerson Method

The Ford-Fulkerson method is iterative. We start with $f(u, v) = 0$ for all u, v ∈ V, giving an initial flow of value 0. Each iteration, we increase the flow value by finding an augmenting path, which we can think of simply as a path from the source s to the sink t along which we can send more flow, and then augmenting the flow along this path. We repeat this process until no augmenting path can be found.

Each iteration of the Ford-Fulkerson method, we find some augmenting path $p$ and increase the flow $f$ on each edge of $p$ by the residual capacity $c_f(p)$. The residual capacity $c_f(u, v)$ is computed in accordance with the formula (1). The expression $c_f(p)$ in the code is actually just a temporary variable that stores the residual capacity of path p.

The Ford-Fulkerson algorithm simply expands on the Ford-Fulkerson method pseudo code given earlier. In lines 4–8 of the while loop, repeatedly finds an augmenting path $p$ in $G_f$ and augments flow $f$ along $p$ by the residual capacity $c_f(p)$. When no augmenting paths exist, the flow $f$ is a maximum flow.

#### f) Finding A Maximum Bipartite Matching

We can use the Ford-Fulkerson method to find a maximum matching in a bipartite graph G= (V, E) in time polynomial in |V| and |E|. The trick is to construct a flow network in which flows correspond to matching, we define the corresponding flow network G'= (V', E') for the bipartite graph G as follows. We let the source s and sink t be new vertices not in V, and we let V'=V∪{s,t}. If the vertex partition of G is V = L ∪ R, the directed edges of G' are the edges of E, directed from L to R, along with V new edges:

$$E'=\{(u,v):u{\in}L,v{\in}R,and(u,v){\in}E\}\,U\{(s,u):u{\in}L\}\,U\{(v,t):v{\in}R\} \quad (4)$$

In "fig. 1", we have shown a bipartite graph which is made up of output parameters of requested (R) and advertised (Q) web services and augmenting path is shown by dark lines.

By applying Ford Fulkerson algorithm on this bipartite graph, we can compute the matching rate between two web services, R and Q. we should make such a graph for these two services' inputs ($R_{in}$, $Q_{in}$). We consider the average of input similarity rate and output similarity rate for two web services similarity rate.

### C. *Selecting The Most Similar Web Service*

We choose the first service as the best one, and do the first and second steps, for each service. If the next service has higher similarity rate, we choose it as the best and so on. If one service has the similarity rate 10 (which is the maximum possible) the algorithm finishes and we don't check the other services.

---

**ALGORITHM 1 : FORD-FULKERSON-METHOD (G, s, t)**

*1: initialize flow f to 0*
*2: while there exists an augmenting path p*
*3:    do augment flow f along p*
*4: return f*

**ALGORITHM 2 : FORD-FULKERSON (G, s, t)**

*1:for each edge (u, v) ∈ E[G]*
*2:   do f [u, v] ← 0*
*3:      f [v, u] ← 0*
*4:while there exists a path p from s to t in the residual network $G_f$*
*5:   do $c_f (p)$ ← min {$c_f (u, v) : (u, v)$ is in p}*
*6:      for each edge (u, v) in p*
*7:         do f [u, v] ← f [u, v] + $c_f(p)$*
*8:            f [v, u] ← − f [u, v]*

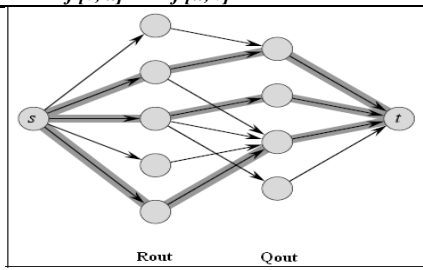

Figure 1.   Bipartite matching graph

### IV. DISCUSSION

The running time of Ford-Fulkerson depends on how the augmenting path *p* in line 4 of algorithm 2 is determined. The **while** loop of lines 4–8 is executed at most | $f_{max}$| times, since the flow value increases by at least one unit in each iteration. The work done within the **while** loop can be made

efficient if we efficiently manage the data structure used to implement the network G = (V, E). Let us assume that we keep a data structure corresponding to a directed graph G' =(V, E'), where E'={(u, v): (u, v)∈ E or (v, u)∈ E}. Edges in the network G are also edges in G', and it is therefore a simple matter to maintain capacities and flows in this data structure.

Given a flow *f* on G, the edges in the residual network $G_f$ consist of all edges (u, v) of G' such that $c (u, v) − f [u, v]$ ≠ 0. The time to find a path in a residual network is therefore O (V + E') = O (E) if we use either depth-first search or breadth-first search. Each iteration of the **while** loop thus takes O(E) time, making the total running time of Ford-Fulkerson O(E | $f_{max}$|) where $f_{max}$ is the maximum flow found by the algorithm.

In table 2, we have compared the Ford-Fulkerson algorithm with the other algorithms. As you see, Ford-Fulkerson algorithm has the best (less) running time among all algorithms that are used to solve bipartite matching algorithms [14].

### V. CONCLUSION AND FUTURE WORK

In this paper we have identified the problem of discovering a semantic web and proposed a new method to solve this problem. By the help of UDDI registry, which advertises services that are all described in OWL language, we could compare the functionality of a requested service with others. We wrote some rules for calculating for calculating the substitution rate of two parameters.

TABLE II.    DIFFERENT ALGORITHMS FOR SOLVING MAXIMUM FLOW PROBLEM

| Name of algorithm | Running time | Description |
|---|---|---|
| Ford-Fulkerson | O(E|$f_{max}$|) | The algorithm works only if all weights are integers. Otherwise it is possible that the Ford-Fulkerson algorithm will not converge to the maximum value. |
| Edmonds-Karp | O(V.$E^2$) | A specialization of Ford-Fulkerson, finding augmenting paths with breadth-first search. |
| Relabel-to-Front | O($V^3$) | In each phase the algorithms builds a layered graph with breadth-first search on the residual graph. The maximum flow in a layered graph can be calculated in O (VE) time, and the maximum number of the phases is n-1. In networks with unit capacities, Dinic's algorithm terminates in time. |
| Push-Relabel | O($V^2$.E) | It is the extension of Relabel-to-Front algorithm. |

We used the bipartite graph for calculating the matching rate of two semantic web services in our work. Then, by the help of Ford-Fulkerson method, we could compute this similarity rate. By comparing Ford-Fulkerson method with other algorithms, we proofed that the best algorithm for computing semantic web matching is the one we had used, Ford-Fulkerson. This is our innovation, because no one had used Ford-Fulkerson method for semantic web service matching, yet.

Some of web services have preconditions to do a work and then have some effects. By considering these two factors, we can obtain a better result, certainly.

Our future work is focused on improving the efficiency and accuracy of this method by considering preconditions and effects of a service.